\pdfoutput=1

\documentclass[11pt]{article}

\usepackage[final]{acl}

\usepackage{times}
\usepackage{latexsym}
\usepackage[T1]{fontenc}
\usepackage[utf8]{inputenc}
\usepackage{microtype}
\usepackage{inconsolata}
\usepackage{graphicx}
\usepackage{booktabs}
\usepackage{pgfplots}
\usepackage{subcaption}
\usepackage{amsmath}
\usepackage{amssymb}
\usepackage{xcolor}
\usepackage{colortbl}
\usetikzlibrary{patterns}
\pgfplotsset{compat=1.18}
\definecolor{lightgray}{gray}{0.95}
\usepackage{listings}

\newcommand{\methodname}{Reasoning Paths Optimization}
\newcommand{\declarelogo}[0]{\includegraphics[height=.02\textwidth]{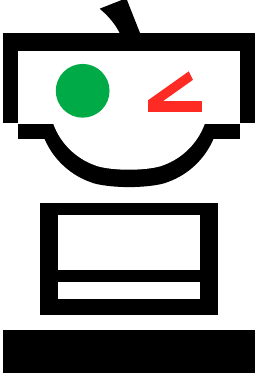}}

\newcommand{\flag}[1]{{ #1}}

\title{\methodname{}:\\
\flag{Learning to Reason and Explore From Diverse Paths}}

\author{
\textbf{
Yew Ken Chia\thanks{~~Equal contribution. Yew Ken and Guizhen are students under the Joint PhD Program between Alibaba and their corresponding university. 
}
~\textsuperscript{\rm 1,~${\declarelogo}$}
\quad
Guizhen Chen\footnotemark[1]\textsuperscript{\rm ~~~1,~2}
\quad Weiwen Xu\thanks{~~Corresponding author.}\textsuperscript{\rm ~1}
} \\
\textbf{
Luu Anh Tuan\textsuperscript{\rm ~2} 
\quad
Soujanya Poria\textsuperscript{\rm ${\declarelogo}$}
\quad 
Lidong Bing\textsuperscript{\rm ~1}
} \\
\textsuperscript{\rm ${\declarelogo}$} Singapore University of Technology and Design ~~\\
\textsuperscript{\rm 1}DAMO Academy, Alibaba Group, Singapore~~  \\
\textsuperscript{\rm 2}Nanyang Technological University, Singapore \\
{\tt\{yewken\_chia, sporia\}@sutd.edu.sg} 
\quad
{\tt\{guizhen001, anhtuan.luu\}@ntu.edu.sg} \\
{\tt\{yewken.chia, guizhen.chen, xuweiwen.xww, l.bing\}@alibaba-inc.com}
}

\begin{document}
\maketitle

\begin{abstract}

\flag{Advanced models such as OpenAI o1
exhibit impressive problem-solving capabilities through step-by-step reasoning. However, they may still falter on more complex problems, making errors that disrupt their reasoning paths.}
We attribute this to the expansive solution space, where each step has the risk of diverging into mistakes. 
\flag{To enhance language model reasoning,}
we introduce a specialized training framework called \methodname{} (RPO), which 
\flag{enables learning to reason and explore from diverse paths.}
Our approach encourages favorable branches at each reasoning step while penalizing unfavorable ones, enhancing the model's overall problem-solving performance. 
\methodname{} does not rely on large-scale human-annotated rationales or outputs from closed-source models, making it scalable and data-efficient. We focus on multi-step reasoning tasks, such as math word problems and science-based exam questions.
The experiments demonstrate that our framework significantly enhances the reasoning performance of large language models, with up to 3.1\% and 4.3\% improvement on GSM8K and MMLU (STEM) respectively. 
\flag{Our data and code can be found at \url{https://reasoning-paths.github.io}}.

\end{abstract}

\section{Introduction}

\begin{figure*}[t]
    \centering
    \includegraphics[width=0.95\textwidth]{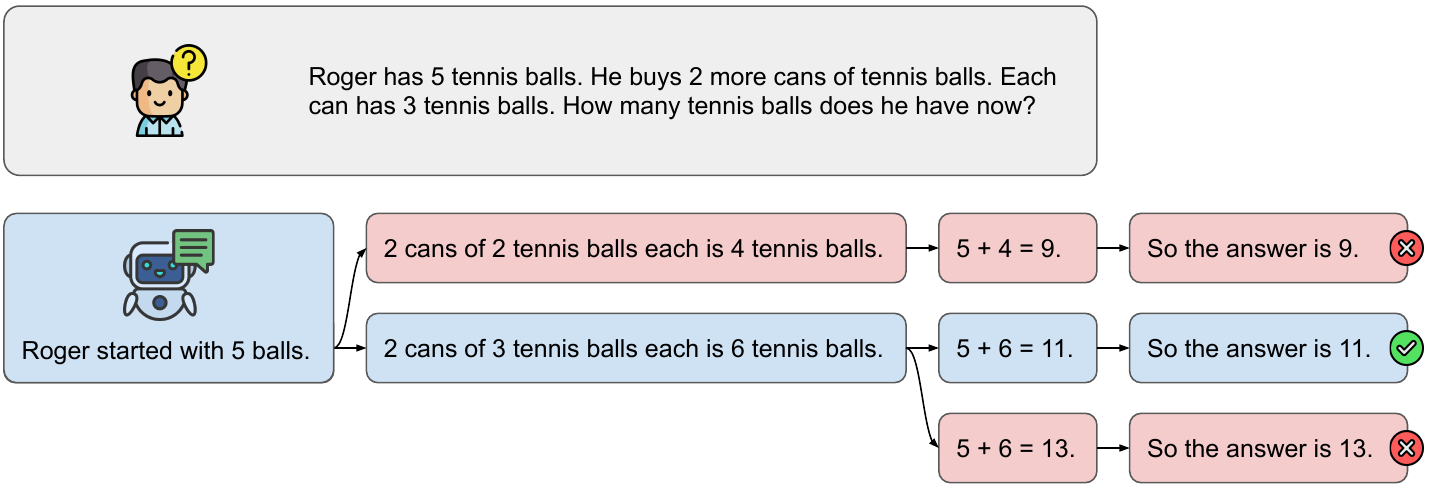}
    \caption{An example of how the reasoning path of the model can easily diverge to unfavorable branches that fail to reach the correct solution. While we show a simplified example here, the challenge is amplified for more complex questions that require longer reasoning paths.
    }
    \label{fig:intro}
\end{figure*}

Large language models (LLMs) have shown remarkable proficiency in following instructions and reasoning \citep{brown2020language, ouyang2022training, touvron2023llama, jiang2023mistral}. Analogous to human cognitive processes, chain-of-thought prompting guides models to reason step-by-step before producing the final answer \citep{wei2022chain}, significantly boosting their reasoning capabilities and demonstrating exceptional performance across a wide array of tasks \citep{wang2023boosting,Chung2024Scaling}.
Despite these advancements, LLMs still exhibit limitations in scenarios that require 
more complex reasoning \cite{zhong2024evaluationopenaio1opportunities}.

As shown in Figure \ref{fig:intro},
the step-by-step reasoning path of the model is at risk of diverging to unfavorable branches that contain mistakes, thus reducing the chance of reaching the correct solution.
While such mistakes may not immediately lead to the wrong answer, they can compound and derail the reasoning process \cite{ling2023deductiveverify}.
Furthermore, this challenge is amplified for more complex problems such as competition-level math questions \cite{hendrycks2021measuringmath} that require long reasoning paths to solve.
Hence, there is a need to address this challenge by encouraging the models to generate the correct reasoning path while avoiding the unfavorable branches.

To ensure a trustworthy answer derivation process, prior studies have explored a range of methods, encompassing both prompting and fine-tuning techniques.
Prompting methods repeatedly sample from LLMs for the same question and employ a voting mechanism to select the most accurate reasoning step among several alternatives.
Such voting mechanisms can be applied at the final stage of the process, as demonstrated in Self-Consistency \citep{wang2023selfconsistency}, or at every intermediate step, as illustrated in Tree-of-Thought \citep{yao2023tree}.
\citet{yao2023react} shows that leveraging external environmental feedback could remind LLMs of some potential errors within their reasoning process, which potentially prevents these errors from affecting subsequent steps.
However, the prompting methods generally demand extensive token usage to explore multiple reasoning paths from LLMs and integrate feedback from the environment. This causes a significant computational cost and huge execution latency.

\begin{figure*}[t]
    \centering
    \includegraphics[width=\textwidth]{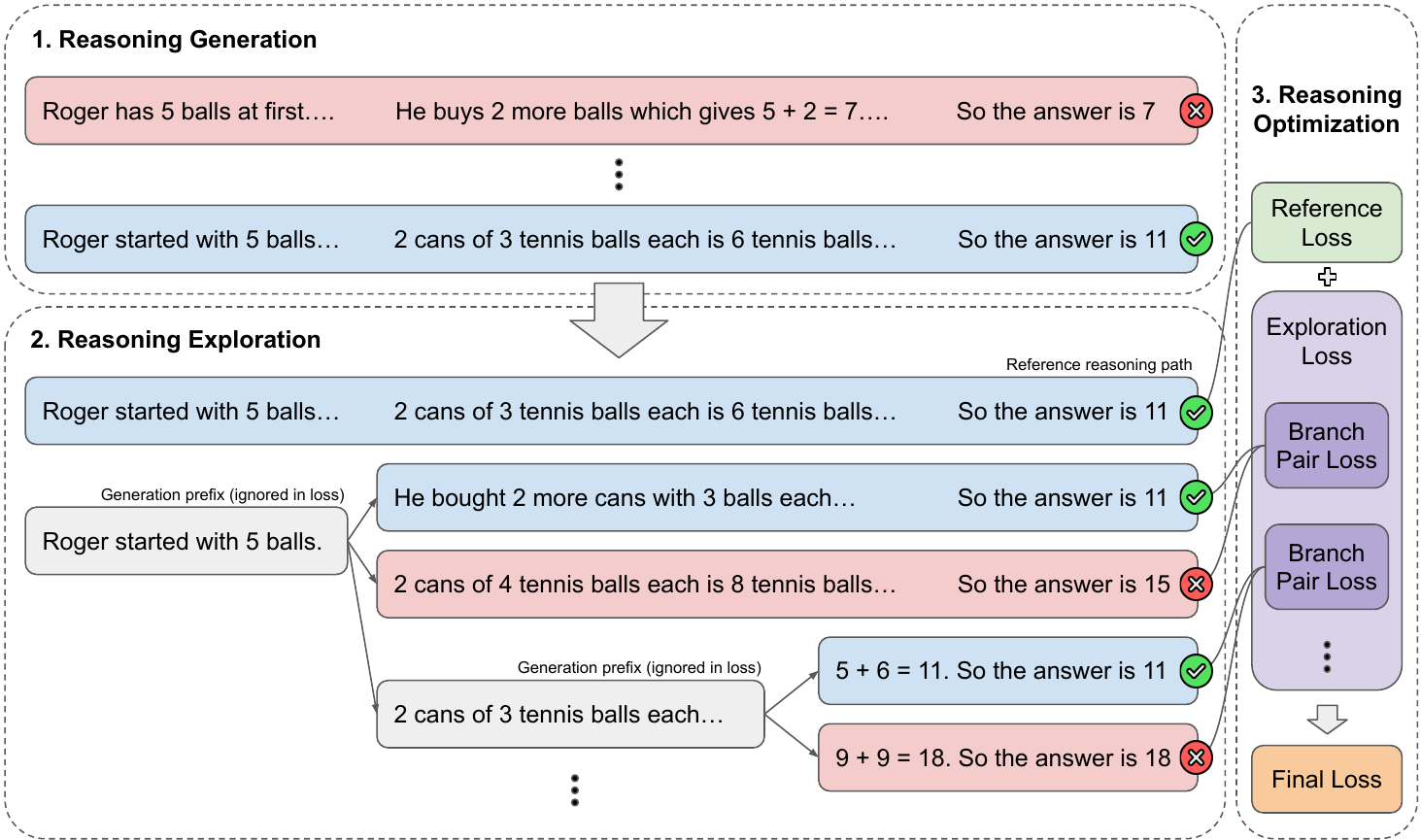}
    \caption{An overview of our \methodname{} framework for exploring and learning over diverse reasoning paths. 
    }
    \label{fig:framework}
\end{figure*}

Alternatively, fine-tuning methods can directly enhance the reasoning capability of LLMs without exhaustive prompting engineering.
Among these methods, reinforcement learning from human feedback (RLHF) \citep{christiano2017deep,stiennon2020learning,ouyang2022training}, which involves training a reward model to optimize LLMs, has shown considerable effectiveness in aligning LLMs. This approach further spurs the development of subsequent works focused on preference optimization, such as DPO \citep{rafailov2023directdpo} and SimPO \citep{meng2024simpo}, which has gained widespread practical adoption due to its simplicity and stability. However, it has been observed that these preference optimization algorithms may be less effective or even detrimental to tasks requiring in-depth reasoning \citep{meng2024simpo}. We hypothesize that these optimization methods may indiscriminately target the entire reasoning path as problematic, whereas, as indicated in Figure \ref{fig:framework}, errors in reasoning often occur at specific steps and affect only the subsequent erroneous branches.



To address the challenge of LLMs committing mistakes that can derail their reasoning paths, we introduce \methodname{}, a novel framework designed to explore and learn from varied reasoning paths. 
As illustrated in Figure \ref{fig:framework}, our approach initiates by generating a reference reasoning path for each question that can reach the correct answer via chain-of-thought prompting. Following this, we explore various solution branches emanating from each step in the reference path. With the reference reasoning paths and the potential solution branches explored, we optimize the model from two critical angles: (1) The model should generate the reference reasoning path with a high probability. (2) The model should favor all potential branches leading to the correct answer over those that do not. 
To achieve the optimization, we propose a reference loss that maximizes the likelihood of generating the reference reasoning path and an exploration loss that provides contrastive feedback over each pair of favorable and unfavorable branches.
As a result, we can explore the diverse mistakes the model is liable to produce, and reduce their occurrence by aligning the models to the correct reasoning path. 

Experimental results on math-based reasoning tasks such as GSM8K \cite{cobbe2021traininggsm8k} and MATH \cite{hendrycks2021measuringmath} demonstrate the effectiveness of our approach compared to strong baselines.
In addition, we show that \methodname{} can generalize beyond math tasks to improve reasoning performance on the science, technology, engineering, and math (STEM) subset of the MMLU \cite{hendrycks2021measuringmmlu} exam question dataset.
Notably, the experiments show up to 3.1\% and 4.3\% improvement compared to the high-performing baseline on GSM8K and MMLU (STEM) datasets respectively.

\section{\methodname{}}

\subsection{Task Formulation}

In this work, we focus on problems that require multiple steps to arrive at the final answer or produce the final result, such as math word problems \cite{gsm8k, hendrycks2021measuringmath}.
Thus, we provide a concrete task formulation in this section.
Given a question \( Q \) posed in natural text, the goal is to produce the final answer \( A \) in natural text. The model is assumed to go through several reasoning steps \( S_1, S_2, \ldots, S_n \) to arrive at the final answer \( A \). The reasoning path \( P \) is defined as the sequence of these steps:

\begin{align}
P = (S_1, S_2, \ldots, S_n)
\end{align}
where each \( S_i \) is a natural language sentence 
and the last step $S_n$ should contain the answer $A$.
Although the model could generate the correct final answer either devoid of a reasoning path or containing faulty reasoning steps, the findings and analysis presented in Section \ref{sec:experiments} suggest that such an outcome is improbable.
\flag{
To determine the separation points between different steps in the generation, we utilize the punctuation-based sentence splitting tool from NLTK, as the reasoning process follows a natural language structure.}

\subsection{Framework Overview}

Large language models are capable of reasoning step-by-step to enhance their problem-solving abilities. However, they often fall short when faced with more challenging problems, committing mistakes that derail their reasoning paths. We believe this issue arises from the large solution space, where multiple reasoning paths can lead to the correct final answer, but each step carries the risk of branching into errors.
To address this, we propose a specialized training framework that jointly considers diverse reasoning paths for a given problem. Our approach encourages favorable branches at each reasoning step while penalizing unfavorable ones. This framework, which we call Reasoning Paths Optimization (RPO), consists of three main stages as shown in Figure \ref{fig:framework}:

\textbf{1. Generation}: 
The generation stage aims to elicit correct reasoning steps from the base model to serve as reference reasoning paths.
This eliminates the need for acquiring ground-truth reasoning path annotations.

\textbf{2. Exploration}: 
To effectively explore the potential solution space to a given problem, this stage progressively creates branches from each step along reference reasoning paths. As a result, we can obtain multiple favorable and unfavorable reasoning branches, which will be used to provide contrastive feedback to the model.

\textbf{3. Optimization}: This final stage aggregates and optimizes according to the reference reasoning paths and explored branches to enhance the innate reasoning ability of the base model. Thus, our framework aims to improve the overall reasoning ability of large language models. 


\subsection{Reasoning Generation}
\label{sec:RG}
While training with explanations or step-by-step reasoning paths \cite{mukherjee2023orca} can improve the reasoning performance of language models, it is labor-intensive and 
costly
to annotate such data.
Hence, our framework begins with
a reasoning generation stage that automatically generates the reference reasoning paths.
Concretely, given a problem question \( Q \), we use chain-of-thought prompting \cite{wei2022chain} to generate reasoning paths. 
The chain-of-thought demonstration input $D_{CoT}$ consists of $m$ ground-truth examples, 
where each example is a pair consisting of a problem question and its corresponding reasoning path.

Let $M$ be the base model, and we sample the reference reasoning path $P$ by prompting the model with the chain-of-thought demonstration $D_{CoT}$ and the given question $Q$. 
We use temperature sampling \cite{fan-etal-2018-hierarchical} with a fixed temperature $T$:

\begin{align}
    P \sim M(D_{CoT}, Q | T)
\end{align}

We consider the generated path as correct if it concludes with a correct answer. Therefore, we define the following function \( \mathcal{F} \) to verify if the last step $S_n \in P$ contains the ground-truth answer $A$:

\begin{align}
\mathcal{F}(P) = 
\begin{cases} 
1 & \text{if } A \in S_n \\
0 & \text{otherwise}
\end{cases}
\end{align}

If the outputs are incorrect, i.e., \( \mathcal{F}(P_i) = 0 \), we repeat the sampling and verification process until \( \mathcal{F}(P_i) = 1 \) 
with a cap of 10 attempts, i.e., $i \leq 10$.
If no suitable path is obtained after multiple attempts, we deem that this problem is far beyond the ability of the model and remove it from the training set.
Thus, we obtain an initial dataset $D_{init}$
which contains the original questions, the ground-truth answers, and model-generated reference paths.


\subsection{Reasoning Exploration}
\label{sec:RE}
To consider potential mistakes that can occur from each reasoning step, the exploration stage of our framework explores multiple branches at each step.
Concretely, given the problem \( Q \), chain-of-thought demonstration, and previous steps of the generated reasoning path \( P_{1:i-1} = (S_1, S_2, \ldots, S_{i-1}) \), we use temperature sampling \cite{fan-etal-2018-hierarchical} to obtain diverse branches from the current point in the reasoning path:

\begin{align}
    B_i \sim M(D_{CoT}, Q, P_{1:i-1} | T),
\end{align}
where each branch $B_i = (S'_{i}, S'_{i+1}, \ldots, S'_l)$ should contain the current step up to the final step. 
We aim to obtain a favorable branch $B_i^+$ and an unfavorable branch $B_i^-$ 
where the favorable branch leads to the correct final answer, and the unfavorable branch does not: 
\begin{align}
\mathcal{F}(B_i^+) = 1, \quad
\mathcal{F}(B_i^-) = 0
\end{align}
To achieve this, we iteratively sample multiple branches starting at each step $S'_i$ and verify each one using the function \(\mathcal{F}\), until we obtain one favorable branch and one unfavorable branch, thus forming a reasoning branch pair $(B_i^+, B_i^-)$.
However, if we are unable to form a branch pair after sampling at most ten branches, the problem is removed from the training set.
This ensures that the training data only includes problems where the model can potentially learn from contrasting between the favorable and unfavorable branches of the reasoning path.

\subsection{Reasoning Optimization}
To optimize the base model $M$, 
we consider both the reference reasoning path $P$ generated in Sec. \ref{sec:RG} and the reasoning branch pairs $(B_i^+, B_i^-)$ explored in Sec \ref{sec:RE}.
Concretely, we encourage the model to produce a higher likelihood over the reference reasoning path. This is achieved by applying standard causal language modeling loss \cite{NIPS2000_728f206causallm} 
on
the reference reasoning path $P$, conditioned on the input question $Q$: 
\begin{align}
    \mathcal{L}_{ref} = -\log Pr_M(P \mid Q)
\end{align}

Regarding the branch pair, the comparison between them may reveal the proper direction that guides the model's optimization. 
Therefore, we define a branch pair loss that provides contrastive feedback between the favorable and unfavorable branches.
To formulate the branch pair loss in our framework, we can leverage preference-based objectives from existing work, such as the direct preference \cite{rafailov2023directdpo} or the odds-ratio objective \cite{hong2024orpo}.
In this work, we mainly focus on the objective proposed by \citet{hong2024orpo} due to its simplicity and empirical effectiveness.
Concretely, the branch pair loss $\mathcal{L}_{bp, i}$ at the $i$-th step can be computed as the log odd-ratio between the favorable branch $B_i^+$ and unfavorable branch $B_i^-$, conditioned on the input question $Q$ and reference path $P$:
\begin{align}
    \mathcal{L}_{bp, i} = \log \frac{\textbf{odds}_M(B_{i}^+ \mid Q, P)}{\textbf{odds}_M(B_{i}^- \mid Q, P)}
    \label{eq:branch_pair}
\end{align}
The odds
of generating a branch can be computed as the ratio between the probability of generating the branch and the probability of not generating it, conditioned on the input question $Q$ and the previous steps $P_{1:i-1}$ of the reference path: 

\begin{align}
\textbf{odds}_M(B_i \mid Q, P) = \frac{Pr_M(B_i \mid Q, P_{1:i-1})}{1 - Pr_M(B_i \mid Q, P_{1:i-1})}
\end{align} 

Thus, we can aggregate the loss over the previously explored branch pairs corresponding to each step in the reasoning path:

\begin{align}
    \mathcal{L}_{exp} = \frac{1}{n - 1} \sum_{i=1}^{n} -\log\sigma(\mathcal{L}_{bp, i})
\end{align}
where there are $n$ steps in the reasoning path. We follow \citet{hong2024orpo} to apply the log-sigmoid function $\log\sigma$ on the log odds-ratio for optimization purposes.
Finally, the overall loss $\mathcal{L}_{RPO}$ in our framework is represented as the combination of the reference path loss $\mathcal{L}_{ref}$ and the exploration loss $\mathcal{L}_{exp}$ which provides contrastive feedback over the explored branch pairs:

\begin{align}
    \mathcal{L}_{RPO} = \mathcal{L}_{ref} + \lambda \cdot \mathcal{L}_{exp}
\end{align}
where $\lambda$ is a hyperparameter weight, which intuitively balances 
between optimizing on the reference reasoning path, and the explored branches.

\flag{
We would like to clarify that we compute the loss only on the output tokens. In this case, the output tokens only consist of the incorrect last part, while the correct prefixes serve as the input tokens, which are excluded from the loss calculation as shown in Figure \ref{fig:framework}. Specifically, the reasoning exploration stage in our framework first collect branch pairs from each step along a reference path, then aggregates the branch pair losses conditioned on the input question and the previous steps of the reference path. Therefore, the common prefix between the favorable and unfavorable branch is excluded in the loss calculation.
}

\section{Experiments}
\label{sec:experiments}

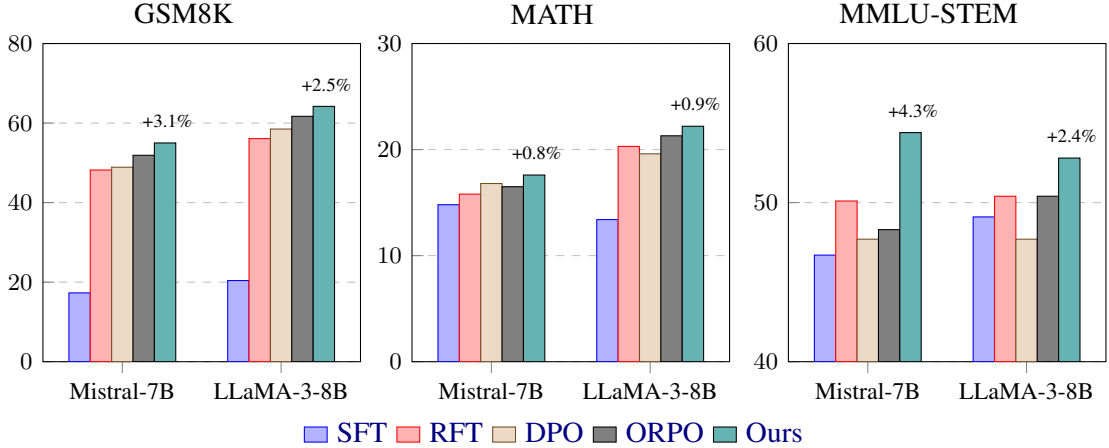
\begin{figure*}[t!]
    \centering
    \begin{subfigure}[b]{0.36\textwidth}
        \centering
        \text{GSM8K} 
        \begin{tikzpicture}
        \begin{axis}[
            width=\textwidth,
            height=5.8cm,
            ybar=0cm,
            bar width=8pt,
            ymax=80,
            ymin=0,
            xtick = {1,2,3,4,5},
            xticklabels = {Mistral-7B, LLaMA-3-8B},
            ticklabel style={font=\small}, 
            xtick pos = left,
            ytick pos = left,
            ymajorgrids = true,
            ytick={0,20,40,60,80,100},
            enlarge x limits=0.5,
            grid style=dashed, 
            legend style={
                legend columns=-1, 
                draw=none,
                column sep=0.05cm,
            },
            legend image code/.code={
              \draw[#1] (0cm,-0.1cm) rectangle (0.3cm,0.1cm);
            }, 
            legend to name={mylegend},
        ]
        \addplot coordinates {(1, 17.3) (2, 20.4)};    
        \addlegendentry{SFT};
        \addplot coordinates {(1, 48.2) (2, 56.1)};    
        \addlegendentry{RFT};
        \addplot coordinates {(1, 48.9) (2, 58.5)};    
        \addlegendentry{DPO};
        \addplot coordinates {(1, 51.9) (2, 61.7)};    
        \addlegendentry{ORPO};
        \addplot[fill=teal!60] coordinates {(1, 55.0) (2, 64.2)};    
        \addlegendentry{Ours};
        \node at (axis cs:1,55.0) [anchor=south, xshift=17pt, yshift=2pt] {\scriptsize +3.1\%};
        \node at (axis cs:2,64.2) [anchor=south, xshift=17pt, yshift=2pt] {\scriptsize +2.5\%};
    \end{axis}
    \end{tikzpicture}
    \label{fig:data_stats}
    \end{subfigure}%
    \hspace{-1.0cm}
    \begin{subfigure}[b]{0.36\textwidth}
        \centering
        \text{MATH} 
        \begin{tikzpicture}
        \begin{axis}[
            width=\textwidth,
            height=5.8cm,
            ybar=0cm,
            bar width=8pt,
            ymax=30,
            ymin=0,
            xtick = {1,2,3,4,5},
            xticklabels = {Mistral-7B, LLaMA-3-8B},
            ticklabel style={font=\small}, 
            xtick pos = left,
            ytick pos = left,
            ymajorgrids = true,
            ytick={0,10,20,30,40,60,80,100},
            enlarge x limits=0.5,
            grid style=dashed, 
        ]
        \addplot coordinates {(1, 14.8) (2, 13.4)};    
        \addplot coordinates {(1, 15.8) (2, 20.3)};    
        \addplot coordinates {(1, 16.8) (2, 19.6)};    
        \addplot coordinates {(1, 16.5) (2, 21.3)};    
        \addplot[fill=teal!60] coordinates {(1, 17.6) (2, 22.2)};  
        
        \node at (axis cs:1,17.6) [anchor=south, xshift=17pt, yshift=2pt] {\scriptsize +0.8\%};
        \node at (axis cs:2,22.2) [anchor=south, xshift=17pt, yshift=2pt] {\scriptsize +0.9\%};
    \end{axis}
    \end{tikzpicture}
    \label{fig:lengths2}
    \end{subfigure}
    \hspace{-1.0cm}
    \begin{subfigure}[b]{0.36\textwidth}
    \centering
    \text{MMLU-STEM} 
    \begin{tikzpicture}
    \begin{axis}[
        width=\textwidth,
        height=5.8cm,
        ybar=0cm,
        bar width=8pt,
        ymax=60,
        ymin=40,
        xtick = {1,2,3,4,5},
        xticklabels = {Mistral-7B, LLaMA-3-8B},
        ticklabel style={font=\small}, 
        xtick pos = left,
        ytick pos = left,
        ymajorgrids = true,
        ytick={0,20,40,50,60,80,100},
        enlarge x limits=0.5,
        grid style=dashed, 
    ]
    \addplot coordinates {(1, 46.7) (2, 49.1)};    
    \addplot coordinates {(1, 50.1) (2, 50.4)};    
    \addplot coordinates {(1, 47.7) (2, 47.7)};    
    \addplot coordinates {(1, 48.3) (2, 50.4)};    
    \addplot[fill=teal!60] coordinates {(1, 54.4) (2, 52.8)};  
    
    \node at (axis cs:1,54.4) [anchor=south, xshift=17pt, yshift=2pt] {\scriptsize +4.3\%};
    \node at (axis cs:2,52.8) [anchor=south, xshift=17pt, yshift=2pt] {\scriptsize +2.4\%};
    \end{axis}
    \end{tikzpicture}
    \label{fig:lengths3}
    \end{subfigure}
    
    \ref{mylegend}
    \caption{Main results showing the evaluation accuracy (\%) of different training methods on math reasoning questions in GSM8K and MATH, and science-based exam questions in MMLU-STEM. We also indicate the improvement of our method compared to the highest-performing baseline. 
    }
    \label{fig:main_results}
\end{figure*}

\subsection{Datasets}
As we focus on enhancing the step-by-step reasoning ability of large language models, we evaluate our approach on datasets of various difficulty levels, including GSM8K \cite{gsm8k} for math word problems and MATH \cite{hendrycks2021measuringmath} for competition-level mathematics.
We use the original training, validation, and testing data splits for our training and evaluation setup.
On the other hand, we also include the MMLU \cite{hendrycks2021measuringmmlu} exam question dataset to evaluate the effectiveness of our approach in other domains.
However, as many of the exam questions focus on world-knowledge and do not require multi-step reasoning, we extract a subset covering 3375 questions in the science, technology, engineering, and math (STEM) domains, and denote this as the MMLU-STEM dataset.
The dataset details can be found in Appendix \ref{sec:data_details}.

Note that our \methodname framework does not necessitate large-scale annotated reasoning paths for training LLMs. On the contrary, for each task, we only need a small number of reasoning demonstrations for implementing CoT prompting, which is easy to obtain. Specifically, we randomly select four questions from the training data and use their ground-truth reasoning path as CoT demonstrations during the reasoning generation stage. For the remaining procedure, \methodname only involves the ground-truth answer to verify the correctness of the explored branch.
We include the prompt examples in Appendix \ref{sec:prompting}.



\subsection{Implementations}
\label{sec:setup}

To evaluate our approach, we implement Mistral-7B and LLaMA-3-8B as our base models, which are recent and popular foundation large language models in the Mistral \cite{jiang2023mistral} and LLaMA \cite{llama2} model families respectively.
To our knowledge, these are the leading foundation models in this parameter size category at the time of writing.
To investigate how our approach affects models of different training stages, we also include experiments show that our framework also benefits the LLaMA-3-8B-Instruct version in Appendix \ref{sec:instruct_experiments}, which has undergone general instruction-tuning \cite{llama2} to enhance performance in many aspects.
Due to computational resource constraints, we are unfortunately unable to train larger model versions such as LLaMA-3-70B in this work.
To avoid potential confounding factors, we do not evaluate on models that already have extensive math-specific training, such as Llemma \cite{azerbayev2024llemma}.
To train the models, we use LoRA fine-tuning \cite{hu2022lora} with a fixed batch size of 8 and a learning rate of 5e-5.
More training details and hyperparameters can be found in the Appendix \ref{sec:training_details}.
To sample multiple outputs from the models, we use a fixed sampling temperature of 0.5.
For evaluation, we use greedy decoding for generation, and the accuracy metric for scoring.

\subsection{Comparison Methods}

To demonstrate the effectiveness of our approach, we compare against strong baselines including reasoning-specific training methods and preference-based optimization methods: 


\begin{enumerate}
    \item Supervised Fine-Tuning (SFT): As a supervised baseline, we consider the case of not using any reasoning paths for training, and only training the model to directly generate the ground-truth final answer.
    \item Rejection Sampling Fine-Tuning (RFT) \cite{yuan2024scaling}: We include RFT as a strong baseline for supervised training, which leverages the model to self-generate reasoning paths for training. We note that this approach is analogous to the reasoning generation stage in our framework, which aims to overcome the data limitation of not having ground-truth reasoning paths.
    \item Direct Preference Optimization (DPO) \cite{rafailov2023directdpo}: As our method contrasts the favorable and unfavorable reasoning branches, it is similar in motivation to DPO which provides the model with contrastive feedback.
    \item Odds-Ratio Preference Optimization (ORPO) \cite{hong2024orpo}: Lastly, we compare against ORPO which proposed the odds ratio objective for preference-based optimization. The main difference between our approach and ORPO is that \methodname{} is a holistic framework specifically designed for reasoning-based tasks; We consider that reasoning mistakes are liable to occur at any step in the reasoning path, and hence explore the possible solution paths which are necessary to provide contrastive feedback over diverse reasoning branch pairs. 
\end{enumerate}

To ensure a fair comparison between different methods, we implement the data setting such that each method uses all viable training samples.
For instance, SFT uses all the training samples as the data setting stipulates that all samples contain the question and ground-truth final answers.
On the other hand, RFT uses only the samples for which the model can generate at least one correct reasoning path, and the preference-based methods DPO and ORPO use only the samples for which the model can generate at least one correct reasoning path and one incorrect reasoning path.
Similar to our approach, the baselines other than SFT use a fixed temperature for sampling reasoning paths with chain-of-thought prompting.
If the model is unable to generate a correct reasoning path after sampling a maximum of ten times, the given question is removed from the training set.

\subsection{Main Results}

To demonstrate the effectiveness of \methodname{}, we compare with strong baselines as shown in Figure \ref{fig:main_results}.
We observe that our approach shows consistent improvements in performance on different datasets and models.
Particularly when trained on top of Mistral-7B, our approach can achieve up to 3.1\% and 4.3\% improvement compared to the highest-performing baseline on GSM8K and MMLU-STEM respectively.
Given that MATH is a relatively difficult task, the base models may struggle to generate the correct paths, thereby limiting the effectiveness of path-based methods. 
Nevertheless, our approach can still improve other baselines, which shows that our approach can more effectively learn from the explored reasoning paths.
On the other hand, we find that SFT performance is lower compared to the other methods trained on self-explored reasoning paths.
This indicates that while it is possible for the model to directly generate the answer without any reasoning steps, it is less effective for more complex reasoning problems.

We further investigate the performance of our method on commonsense and general reasoning tasks in Appendix \ref{sec:csqa}. These tasks typically consist of straightforward questions that do not require lengthy reasoning steps, which may possibly contribute to the high SFT performance. Nevertheless, when the model is prompted to engage in step-by-step reasoning, our framework outperforms other preference optimization approaches, demonstrating its effectiveness in multi-step thinking.

\begin{figure}
\centering
\begin{tikzpicture}
\pgfplotsset{width = \linewidth, height = 5cm}
    \begin{axis}[
        ymax=24,
        ymin=20,
        ylabel={Accuracy (\%)},
        xlabel={Loss Weight $\lambda$},
        xtick = {1,2,3,4,5},
        xticklabels = {0.1, 0.3, 0.5, 0.7, 0.9},
        xtick pos = left,
        ytick pos = left,
        ymajorgrids = true,
        grid style=dashed,
    ]
    \addplot [mark=square, mark size=1.6pt, color=orange] plot coordinates {
    (1, 21.4) (2, 22.2) (3, 21.7) (4, 21.3) (5, 21.1)};
    \end{axis}
\end{tikzpicture}
\caption{The effect of exploration loss weight on the MATH dataset performance for LLaMA-3-8B.}
\label{fig:loss_weight}
\end{figure}
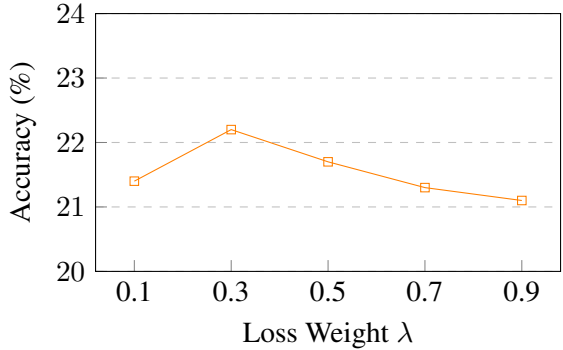

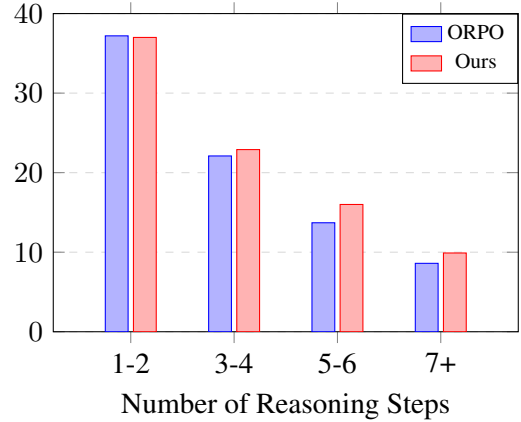
\begin{figure}[t]
\centering
\begin{tikzpicture}
\begin{axis}[
    ybar,
    bar width=.3cm,
    width=\linewidth,
    height=5.8cm,
    enlarge x limits=0.25,
    xlabel={Number of Reasoning Steps},
    symbolic x coords={1-2, 3-4, 5-6, 7+},
    xtick=data,
    ymin=0,
    ymax=40,
    grid=major,
    xmajorgrids=false, 
    grid style={dashed,gray!30},
            legend style={
        font=\fontsize{8}{1}\selectfont, 
        legend style={row sep=-0.0cm},
        at={(1,1)},
    },
    legend image code/.code={
      \draw[#1] (0cm,-0.1cm) rectangle (0.4cm,0.1cm);
    }, 
]
\addplot coordinates {(1-2, 37.2) (3-4, 22.1) (5-6, 13.7) (7+, 8.6)};
\addplot coordinates {(1-2, 37.0) (3-4, 22.9) (5-6, 16.0) (7+, 9.9)};
\legend{ORPO, Ours}
\end{axis}
\end{tikzpicture}
\caption{Performance with respect to reasoning path length on the MATH dataset for LLaMA-3-8B.
}
\label{fig:analysis_lengths}
\end{figure}

\subsection{Effect of Exploration Weight}

To investigate the effect of reasoning exploration within our framework, we conduct an analysis on the loss weight $\lambda$.
Specifically, a lower value of $\lambda$ would place greater emphasis on the supervised loss over the reference path which leads to the correct answer.
On the other hand, a higher value would place greater weight on the explored branches during training, which contrasts between the favorable and unfavorable branches arising from each reasoning step.
As shown in Figure \ref{fig:loss_weight}, we find that having a very low $\lambda$ value leads to sub-optimal results, as it does not place enough emphasis on the reasoning exploration.
On the other hand, we also find that placing too much weight on exploration also does not benefit the training, as it is still necessary to ground the model sufficiently in the reference reasoning path.
Thus, this shows that we need to balance between the optimization of the reference reasoning path and the other possible exploration branches.

\subsection{Analysis On Reasoning Path Length}



To investigate the effectiveness of our approach with respect to the reasoning complexity, we analyze the evaluation accuracy in Figure \ref{fig:analysis_lengths}.
Compared to ORPO \cite{hong2024orpo} which is the highest-performing baseline, we observe benefits from \methodname{} for longer reasoning paths.
This may indicate that our approach can effectively reduce the occurrence of mistakes 
when solving questions that require more complex reasoning.
In future work, we believe that potential directions include more diverse applications and tasks, such as code generation \cite{chen2021codexhumaneval, austin2021programmbpp}, and code-augmented reasoning \cite{pmlr-v202-gao23fpal, li2023chainofcode}.



\begin{table*}[ht]
	\centering
	\resizebox{\linewidth}{!}{
	    \setlength{\tabcolsep}{0mm}{
            \begin{tabular}{p{2.5cm}@{~} @{~}p{20cm}}
            \toprule
            \textbf{Question} & It takes John 5 minutes to get to the first side of the Rubik's cube.  The second and third sides each take twice that long.  The last 2 sides each take half as long as the first, and once you get the 5th side you have all 6 sides.  How long does the Rubik's cube take?\\
            \midrule    
            \textbf{SFT} & 25 \\
            \midrule
            \textbf{RFT} & 5 + 2*5 + 2*5 + 2*5 + 2*5 + 2*5 = 5 + 10 + 10 + 10 + 10 + 10 = 55 minutes. So the answer is \fbox{55} minutes. \\
            \midrule
            \textbf{DPO} & 5 + 10 + 10 + 5 + 5 = 35 minutes. So the answer is \fbox{35} minutes. \\
            \midrule
            \textbf{ORPO} & 5 + 2*5 + 2*5 + 2*5 + 2*5 + 2*5 = 5 + 10 + 10 + 10 + 10 + 10 = 55 minutes. So the answer is \fbox{55} minutes. \\
            \midrule
            \textbf{RPO} & The first side takes 5 minutes. The second and third sides each take 10 minutes. The last 2 sides each take 2.5 minutes. The total time is 5 + 10 + 10 + 2.5 + 2.5 = 30 minutes. So the answer is \fbox{30} minutes. \\
            \bottomrule
            \end{tabular}
        }
    }
    \caption{A qualitative comparison between GSM8K outputs after training LLaMA-3-8B with different methods. }

\label{tab:case1}
\end{table*}

\subsection{Case Study}
To examine the qualitative benefits, Table \ref{tab:case1} shows an example of outputs produced by different methods for the same question. While SFT provides an incorrect and over-simplified answer without showing any working, RFT, DPO, and ORPO directly calculate without explanation and thus make mistakes at the very first step. In contrast, the detailed breakdown of steps shows that RPO not only arrives at the correct answer but does so through a coherent process.

\subsection{Code-Based Reasoning}

\begin{table}[!t]
    \centering
    \resizebox{\linewidth}{!}{
    \begin{tabular}{lccc}
    \toprule
    \textbf{Training Method}
    &\textbf{GSM8K (Text)} 
    &\textbf{GSM8K (Code)} \\
    \midrule
    ORPO & 61.7 & 61.6 \\
    RPO (Ours) & \textbf{64.2} & \textbf{63.4} \\
    \bottomrule
    \end{tabular}
    }
    \caption{Analysis of performance benefits for text-based reasoning as our main setting, and code-based reasoning through python programs. Experiments are conducted using LLaMA-3-8B.}
    \label{tab:code}
\end{table}

\flag{
Beyond reasoning in natural language works such as PAL \cite{pmlr-v202-gao23fpal} have shown that large language models can be prompted to solve reasoning problems with code.
To this end, we have conducted an analysis to show that our framework can also generalize to code-based reasoning. 
Concretely, in our reasoning generation stage, instead of generating text-based reasoning paths, we prompt the model with code demonstrations to generate a python program, which is executed to obtain the output answer. As shown in Table \ref{tab:code}, we find similar benefits for text-based reasoning and code-based reasoning compared to ORPO, which is our strongest baseline.
}

\subsection{Effect of Contrastive Objectives}

\flag{
To demonstrate the robustness of our framework, we have conducted additional experiments using different objectives to contrast between favorable and unfavorable paths.
Specifically, we show that the odds-ratio objective \cite{hong2024orpo} in our branch pair loss can be easily replaced with the direct preference objective \cite{rafailov2023directdpo} for the branch pair loss in Equation \ref{eq:branch_pair}. 
As shown in Table \ref{tab:dpo}, the consistent benefit across different objectives demonstrates that our framework is robust and outperforms the respective baselines.
}

\begin{table}[!t]
    \centering
    \resizebox{1.0\linewidth}{!}{
    \begin{tabular}{lccc}
    \toprule
    \textbf{Method}
    &\textbf{GSM8K} 
    &\textbf{MATH}
    &\textbf{MMLU} \\
    \midrule
    DPO & 58.5 & 19.6 & 47.7 \\
    ORPO & 61.7 & 21.3 & 50.4 \\
    Ours (w/ direct-preference) & 60.7 & 21.4 & 52.8 \\
    Ours (w/ odds-ratio) & 64.2 & 22.2 & 52.8 \\
    \bottomrule
    \end{tabular}
    }
    \caption{Performance comparison on GSM8K, MATH, and MMLU-STEM datasets for different contrastive objectives in our framework using LLaMA-3-8B.}
    \label{tab:dpo}
\end{table}

\subsection{Effect of Reference Paths}

\flag{
In our exploration stage, we use the first reasoning path with the correct answer as the reference path. However, the correct answer can often be achieved via different paths. To analyse the effect of different reference paths, we select a random path with the correct answer after sampling 10 times. Results in Table \ref{tab:reference_path} show that our method remains effective even with this variation, demonstrating its robustness across different reference paths. In addition, we analyse the effect of using more reference paths, eg, three correct reference paths. The results show that our approach can scale to multiple reference paths to further enhance performance.}

\begin{table}[!t]
    \centering
    \resizebox{1.0\linewidth}{!}{
    \begin{tabular}{lc}
    \toprule
    \textbf{Exploration}
    &\textbf{GSM8K} \\
    \midrule
    ORPO & 61.7 \\
    Ours (w/ first correct as reference path) & 64.2 \\
    Ours (w/ random one correct as reference path) & 63.5 \\
    Ours (w/ random three correct as reference paths) & 65.2 \\
    \bottomrule
    \end{tabular}
    }
    \caption{Performance comparison on GSM8K for different reference paths using LLaMA-3-8B.}
    \label{tab:reference_path}
\end{table}

\section{Related Work}

\paragraph{Alignment and Preference-Based Optimization}
Reinforcement learning from human feedback (RLHF) \cite{DBLP:conf/nips/ChristianoLBMLA17, ouyang2022training, xu-etal-2024-reasonsreject} is a popular technique that aligns large language models with human preferences and to follow instructions \cite{ghosal2023flacunaunleashingproblemsolving, chia-etal-2024-instructeval}. During RLHF, a separate reward model is trained to provide scalar value feedback, which is passed to fine-tune LLMs with PPO algorithm \citep{schulman2017proximal,ziegler2019fine}. However, PPO is known to be complex and unstable \citep{zheng2023secrets}, and the multi-stage training of a reward model and a policy model is also challenging  \citep{meng2024simpo}.
Recently, several techniques, including DPO \cite{rafailov2023directdpo, irpo-DBLP:journals/corr/abs-2404-19733}, IPO \cite{IPO}, SimPO \cite{meng2024simpo}, and ORPO \cite{hong2024orpo}, have been proposed to eliminate the need for a reward model, which significantly stabilize and simplify the training process. They make pairwise comparisons between two responses generated by the models and push the model to assign a higher likelihood to the favorable response over the unfavorable one. However, these preference optimization methods indiscriminately compare the two responses in their entirety, overlooking the fact that errors in multi-step reasoning tasks arise only at specific steps and their subsequent branches. In this work, we propose \methodname{} which considers each intermediate step.

\paragraph{Multi-step Reasoning in Language Models}
Large language models are capable of solving reasoning tasks by generating solutions in a step-by-step manner \cite{nye2022show, wei2022chain, kojima2022large, fu2023complexitybased, chu2024navigate}. For example, \citet{wei2022chain} and \citet{kojima2022large} demonstrate that by guiding the model to generate the reasoning steps before generating the final answer, the multi-step reasoning capabilities of LLMs could be effectively elicited, even in multimodal settings \cite{chia-etal-2024-puzzlevqa, zhang2024multimodal}.
However, LLMs are prone to producing errors during the reasoning process, especially for complex multi-step reasoning tasks \cite{li2024chainofknowledge, Chia2023ContrastiveCP}. To mitigate mistakes in the reasoning steps, a straightforward way is to verify the reasoning paths step-by-step. This encourages further investigations on process supervision.
\citet{Uesato2022SolvingMW} and \citet{lightman2024lets} collect human feedback labels for step-level solutions to verify the intermediate steps generated by reasoning models. 
Recent studies \cite{li-etal-2023-making, wang2024mathshepherd, wang2024multistep} construct the step-wise labels automatically to prevent costly human annotations.
These methods focus on training the verifiers (i.e., reward models). In contrast, we apply process supervision to preference optimization methods, without requiring a separate reward model.

\paragraph{Path Exploration in Artificial Intelligence}
The exploration of diverse paths has been widely used to improve the performance of complex tasks in the field of artificial intelligence.
AlphaGo \cite{alphago} uses Monte Carlo Tree Search \cite{mcts} to explore a large space of possible moves. 
Similarly, \citet{yao2023tree} leverage Tree-of-Thought prompting to explore possible solution space from LLMs.  Other works \cite{feng2023alphazerolike, xie2023selfevaluation} also design tree-based decoding strategies to search for the optimal solution.
In the area of reasoning tasks, previous works have explored using self-sampled solutions for training \cite{ni2023learning} and tree search for path generation \cite{golovneva2023pathfinderguidedsearchmultistep}.
Inspired by these works, we explore the diverse solution space generated by language models. Furthermore, we optimize the models with contrastive feedback from both favorable and unfavorable branches during training.
Inspired by these works, we explore the diverse solution space generated by the models. Furthermore, we optimize LLMs with both favorable and unfavorable branches during training.

\section{Conclusion}
In this paper, we introduced a novel training framework called Reasoning Paths Optimization (RPO) to enhance the step-by-step reasoning capabilities of LLMs. Our approach addresses the challenge of complex problem-solving tasks, where each reasoning step carries the risk of diverging into errors.
RPO considers diverse reasoning branch pairs and encourages favorable branches at each reasoning step while penalizing unfavorable ones.
Our framework is scalable, as it does not rely on large-scale human-annotated rationales. Instead, it leverages the model's own generated reasoning paths, making it adaptable to multi-step reasoning tasks such as math word problems. Through extensive experiments on datasets of varying difficulties, our framework provides an effective approach to enhance reasoning, paving the way for more reliable and accurate problem-solving in complex scenarios. 

\section*{Acknowledgment}

\flag{This work was substantially supported by DAMO Academy through DAMO Academy Research Intern Program.}

\section*{Limitations}
Our framework relies on the model's ability to generate correct reasoning paths during the training phase. If the base model is significantly under-performing, it may struggle to generate the necessary correct paths, thereby limiting the effectiveness of our approach.
To provide performance insights beyond accuracy, we also report the Inter. F1 metric in Appendix \ref{sec:inter_f1}, which demonstrates that the reasoning paths generated after training with our method is more consistent with the ground-truth reasoning paths in GSM8K.
Although the process of generating and exploring multiple reasoning paths for each problem is more computationally intensive, we note that this is a one-time cost during training.
Hence, we believe that this is a worthwhile trade-off to enhance performance, which can be amortized over many inference cases.

\bibliography{custom}
\appendix

\section{Appendix}
\label{sec:appendix}

\begin{figure*}[t!]
    \centering
    \begin{subfigure}[b]{0.5\textwidth}
        \centering
        \text{GSM8K} 
        \begin{tikzpicture}
        \begin{axis}[
            width=\textwidth,
            height=5.8cm,
            ybar=0cm,
            bar width=8pt,
            ymax=100,
            ymin=0,
            xtick = {1,2,3,4,5},
            xticklabels = {LLaMA-3-8B, LLaMA-3-8B-Instruct},
            xtick pos = left,
            ytick pos = left,
            ymajorgrids = true,
            ytick={0,20,40,60,80,100},
            enlarge x limits=0.5,
            grid style=dashed, 
            legend style={
                legend columns=-1, 
                draw=none,
                column sep=0.05cm,
            },
            legend image code/.code={
              \draw[#1] (0cm,-0.1cm) rectangle (0.3cm,0.1cm);
            }, 
            legend to name={mylegend2},
        ]
        \addplot coordinates {(1, 20.4) (2, 21.4)};    
        \addlegendentry{SFT};
        \addplot coordinates {(1, 56.1) (2, 76.8)};    
        \addlegendentry{RFT};
        \addplot coordinates {(1, 58.5) (2, 74.5)};    
        \addlegendentry{DPO};
        \addplot coordinates {(1, 61.7) (2, 77.4)};    
        \addlegendentry{ORPO};
        \addplot[fill=teal!60] coordinates {(1, 64.2) (2, 79.5)};    
        \addlegendentry{Ours};

        \node at (axis cs:1,64.2) [anchor=south, xshift=17pt, yshift=2pt] {\scriptsize +2.5\%};
        \node at (axis cs:2,79.5) [anchor=south, xshift=17pt, yshift=2pt] {\scriptsize +2.1\%};
    \end{axis}
    \end{tikzpicture}
    \label{fig:data_stats2}
    \end{subfigure}%
    \hfill
    \begin{subfigure}[b]{0.5\textwidth}
        \centering
        \text{MATH} 
        \begin{tikzpicture}
        \begin{axis}[
            width=\textwidth,
            height=5.8cm,
            ybar=0cm,
            bar width=8pt,
            ymax=40,
            ymin=0,
            xtick = {1,2,3,4,5},
            xticklabels = {LLaMA-3-8B, LLaMA-3-8B-Instruct},
            xtick pos = left,
            ytick pos = left,
            ymajorgrids = true,
            ytick={0,20,40,60,80,100},
            enlarge x limits=0.5,
            grid style=dashed, 
        ]
        \addplot coordinates {(1, 13.4) (2, 12.4)};    
        \addplot coordinates {(1, 20.3) (2, 29.5)};    
        \addplot coordinates {(1, 19.6) (2, 24.9)};    
        \addplot coordinates {(1, 21.3) (2, 30.0)};    
        \addplot[fill=teal!60] coordinates {(1, 22.2) (2, 31.5)};  
        
        \node at (axis cs:1,22.2) [anchor=south, xshift=17pt, yshift=2pt] {\scriptsize +0.9\%};
        \node at (axis cs:2,31.5) [anchor=south, xshift=17pt, yshift=2pt] {\scriptsize +1.5\%};
    \end{axis}
    \end{tikzpicture}
    \label{fig:lengths22}
    \end{subfigure}
    
    \ref{mylegend}
    \caption{Main results showing the evaluation accuracy (\%) of different training methods on math reasoning datasets. We also indicate the improvement of our method compared to the highest-performing baseline. 
    }
    \label{fig:main_instruct}
\end{figure*}
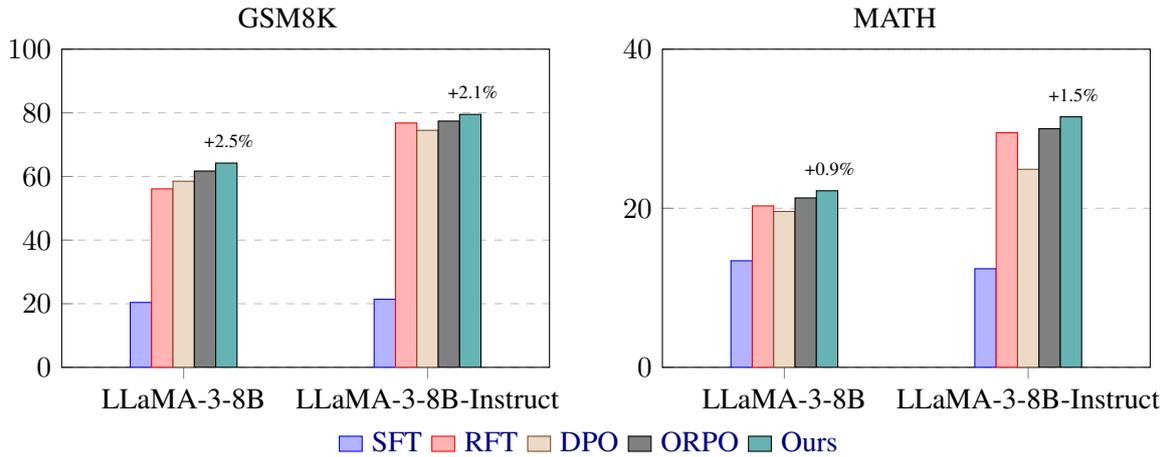

\begin{figure}[t!]
    \centering
    \begin{subfigure}[b]{0.5\textwidth}
        \centering
        \begin{tikzpicture}
        \begin{axis}[
            width=\textwidth,
            height=5.8cm,
            ybar=0cm,
            bar width=8pt,
            ymax=65,
            ymin=40,
            xtick = {1,2,3,4,5},
            xticklabels = {LLaMA-3-8B, LLaMA-3-8B-Instruct},
            xtick pos = left,
            ytick pos = left,
            ymajorgrids = true,
            ytick={0,20,40,50,60,80,100},
            enlarge x limits=0.5,
            grid style=dashed, 
        ]
        \addplot coordinates {(1, 49.1) (2, 53.9)};    
        \addplot coordinates {(1, 50.4) (2, 56)};    
        \addplot coordinates {(1, 47.7) (2, 57.6)};    
        \addplot coordinates {(1, 50.4) (2, 57.3)};    
        \addplot[fill=teal!60] coordinates {(1, 52.8) (2, 58.1)};  
        
        \node at (axis cs:1,52.8) [anchor=south, xshift=17pt, yshift=2pt] {\scriptsize +2.4\%};
        \node at (axis cs:2,58.1) [anchor=south, xshift=17pt, yshift=2pt] {\scriptsize +0.5\%};
    \end{axis}
    \end{tikzpicture}
    \label{fig:lengths_instruct}
    \end{subfigure}
    
    \ref{mylegend}
    \caption{Additional results showing the evaluation accuracy on science, technology, engineering, and math questions in MMLU \cite{hendrycks2021measuringmmlu}. We also indicate the improvement of our method compared to the highest-performing baseline. 
    }
    \label{fig:mmlu_instruct}
\end{figure}
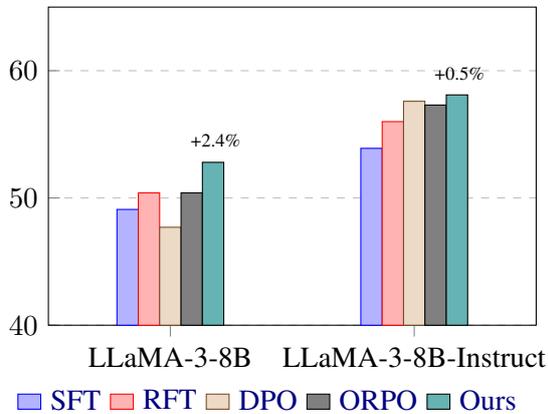

\subsection{Training and Hyperparameter Details}
\label{sec:training_details}


We list the hyperparameter and training details in Table \ref{tab:params}.
Note that we use the validation accuracy of MATH with LLaMA-3-8B to select the loss weight $\lambda \in \{0.1, 0.3, 0.5, 0.7, 0.9\}$ and use it for all datasets.
To ensure the diversity in our reasoning exploration stage, we remove duplicate reasoning paths and branches before training.
To maintain fairness between different training methods, we ensure that each method uses at most one accepted or reference reasoning path that reaches the correct answer for each question. 
For preference-based methods, we ensure that each method uses at most one accepted reasoning path, and one rejected reasoning path that reaches the wrong answer for each question.
Similarly, our approach uses at most one reference reasoning path for each question.
To be fair to DPO which typically follows an SFT training stage, we include the SFT loss over the accepted reasoning path during training, which is a joint loss with the DPO objective.
Hence, all methods in this paper utilize a single training stage.

\flag{
As shown in Table \ref{tab:training_data_sizes}, all training methods use a similar amount of training data in terms of reasoning paths.
Note that the number of samples used for each model is different as the samples are filtered based on the correctness of model outputs. }

\subsection{Dataset Details}
\label{sec:data_details}

For GSM8K and MATH, we use the original training and testing splits. 
For MMLU (STEM), we specifically use the STEM subset for the following subcategories of questions:

\begin{itemize}
    \item abstract\_algebra
    \item astronomy
    \item college\_biology
    \item college\_chemistry
    \item college\_computer\_science
    \item college\_mathematics
    \item college\_physics
    \item computer\_security
    \item conceptual\_physics
    \item electrical\_engineering
    \item elementary\_mathematics
    \item high\_school\_biology
    \item high\_school\_chemistry
    \item high\_school\_computer\_science
    \item high\_school\_mathematics
    \item high\_school\_physics
    \item high\_school\_statistics
    \item machine\_learning
\end{itemize}

We thus create a train-test split of the STEM questions, contain 3000 training and 375 testing samples.

\begin{table}[!t]
    \centering
    \resizebox{0.8\linewidth}{!}{
    \begin{tabular}{lccccccc}
    \toprule
    Loss weight $\lambda$ & 0.3 \\
    Lora rank & 8 \\
    Learning rate & 5e-5 \\
    Batch size & 8 \\
    Training epochs & 3 \\
    Hardware & Single A800 (80GB) \\
    \bottomrule
    \end{tabular}
    }
    \caption{Hyperparameter and training details.}
    \label{tab:params}
\end{table}

\begin{table}[!t]
    \centering
    \resizebox{1.0\linewidth}{!}{
    \begin{tabular}{lcc}
    \toprule
    \textbf{Method}
    &\textbf{LLaMA-3-8B} 
    &\textbf{Mistral} \\
    \midrule
    SFT & 7473 answers & 7473 answers \\
    RFT & 6417 reasoning paths & 5922 reasoning paths \\
    DPO & 5667 reasoning path pairs & 5535 reasoning path pairs \\
    ORPO & 5667 reasoning path pairs & 5535 reasoning path pairs \\
    RPO & 5752 reference paths & 5600 reference paths \\
    \bottomrule
    \end{tabular}
    }
    \caption{Training data comparison for different methods on GSM8K dataset using different models.}
    \label{tab:training_data_sizes}
\end{table}

\subsection{Prompting}
\label{sec:prompting}

For our reasoning generation stage as well as the baselines of RFT, DPO, and ORPO, we use chain-of-thought prompting to generate training reasoning paths. Note that we use 4-shot prompting for all settings and methods as shown below:

\paragraph{GSM8K CoT Prompt}

\begin{verbatim}
Question: There are 180 days in a school 
year.  A senior can skip their final exams
if they miss 5% or less of the school year.
Hazel has missed 6 days of school due to 
illness.  How many more days can she miss 
and still not have to take her exams?

Answer: There are 180 days in the school 
year and she can miss up to 5% so that's 
180*.05 = 9 days\nHazel has been sick 6 
days already and she can only miss 9 days 
or less so she can miss 9-6 = 3 more days.
So the answer is \\boxed{3} days.

Question: Several birds were sitting in 
the branches of a crape myrtle tree. 
There were three times more cardinals
than bluebirds, but half as many swallows
as bluebirds. If there were 2 swallows, 
what is the total number of birds in the 
crape myrtle tree?

Answer: With half as many swallows as 
bluebirds, there are 2*2=4 bluebirds.
With three times more cardinals than 
bluebirds, there are 3*4=12 cardinals,
If there were 2 swallows, then the total 
number of birds in the crape myrtle tree
is 2+4+12=18 birds. So the answer is 
\\boxed{18}.

Question: Barry goes to a shop to buy a
shirt he'd been admiring for quite some 
time. He tells the attendant that it's his
birthday so she decides to give him a 15%
special discount. The price tag on the 
shirt says $80. How much is he supposed 
to pay now, considering the special 
discount?

Answer: 15% of $80 = (15/100)*$80 = $12
The dollar amount of the discount is $12
so he is supposed to pay just $80-$12 = 
$68. So the answer is \\boxed{$68}.

Question: Nancy wanted to make peanut 
butter cookies for a family gathering,
but her cousin is allergic to peanuts. 
She decided to make almond butter cookies
instead. A jar of almond butter costs 
three times the amount that a jar of 
peanut butter does. It takes half a jar 
to make a batch of cookies. A jar of 
peanut butter costs $3. How many dollars
more does it cost per batch to make 
almond butter cookies instead of peanut 
butter cookies?

Answer: A jar of almond butter costs 3 *
3 = $9.\nIt takes half a jar to make a 
batch of cookies, so it costs 9 / 2 = 
$4.50 to use almond butter.\nIt costs 3 
/ 2 = $1.50 to use peanut butter.\nThus,
it costs 4.50 - 1.50 = $3 more to make a
batch of almond butter cookies than 
peanut butter cookies. So the answer is 
\\boxed{$3}.
\end{verbatim}

\begin{table}[!t]
    \centering
    \resizebox{1.0\linewidth}{!}{
    \begin{tabular}{lccccccc}
    \toprule
    \textbf{Method}
    &\textbf{Precision} 
    &\textbf{Recall} 
    &\textbf{F1} \\
    \midrule
    RFT &  81.9 & 75.3 & 77.5 \\
    DPO &  80.8 & 78.9 & 79.0 \\
    ORPO & 82.7 & 78.0 & 79.5 \\
    RPO (Ours) & 83.5 & 79.9 & 80.9 \\
    \bottomrule
    \end{tabular}
    }
    \caption{Analysis of the reasoning quality through Inter. Precision, Recall, and F1 metrics on GSM8K.}
    \label{tab:inter_f1}
\end{table}

\paragraph{MATH CoT Prompt}

\begin{verbatim}
Question: Find the domain of the expression 
$\\frac{\sqrt{x-2}}{\sqrt{5-x}}$.}

Answer: The expressions inside each square 
root must be non-negative. Therefore, $x-2
\ge 0$, so $x\ge2$, and $5 - x \ge 0$, so $x
\le 5$. Also, the denominator cannot be 
equal to zero, so $5-x>0$, which gives 
$x<5$. Therefore, the domain of the 
expression is $\\boxed{[2,5)}$.\nSo the 
final answer is \\boxed{[2,5)}.

Question: If $\det \mathbf{A} = 2$ and 
$\det \mathbf{B} = 12,$ then find $\det 
(\mathbf{A} \mathbf{B}).$

Answer: We have that $\det (\mathbf{A} 
\mathbf{B}) = (\det \mathbf{A})(\det 
\mathbf{B}) = (2)(12) = \\boxed{24}.$
So the final answer is \\boxed{24}.

Question: Terrell usually lifts two 
20-pound weights 12 times. If he uses two
15-pound weights instead, how many times 
must Terrell lift them in order to lift 
the same total weight?

Answer: If Terrell lifts two 20-pound 
weights 12 times, he lifts a total of 
$2\cdot 12\cdot20=480$ pounds of weight.
If he lifts two 15-pound weights instead 
for $n$ times, he will lift a total of 
$2\cdot15\cdot n=30n$ pounds of weight. 
Equating this to 480 pounds, we can solve 
for $n$:\n\\begin{align*}\n30n&=480\\\\ 
\Rightarrow\qquad n&=480/30=\\boxed{16}
\end{align*} So the final answer is
\\boxed{16}.

Question: If the system of equations
\n\\begin{align*}\n6x-4y&=a,\\\\
6y-9x &=b.\n\end{align*}\nhas a solution 
$(x, y)$ where $x$ and $y$ are both 
nonzero, find $\\frac{a}{b},$ assuming 
$b$ is nonzero.

Answer: If we multiply the first equation 
by $-\\frac{3}{2}$, we obtain $$6y-9x=-
\\frac{3}{2}a.$$\nSince we also know that
$6y-9x=b$, we have $$-\\frac{3}
{2}a=b\Rightarrow\\frac{a}{b}=
\\boxed{-\\frac{2}{3}}.$$
So the final answer is 
\\boxed{-\\frac{2}{3}}.
\end{verbatim}

\paragraph{MMLU (STEM Subset) CoT Prompt}

\begin{verbatim}
Question: Which of the following adaptive
features would least likely be found in 
an animal living in a hot arid environment?
\n(A) Long loops of Henle to maximize water
reabsorption\n(B) Storage of water in fatty
tissues\n(C) Large ears to aid in heat
dispersion\n(D) Short loops of Henle to
maximize water secretion

Answer: To determine which adaptive 
feature would be least likely found in
an animal living in a hot arid 
environment, let's analyze each option 
in the context of such an environment: 
(A) Long loops of Henle to maximize 
water reabsorption: In a hot arid 
environment, water is scarce. Animals 
would benefit from adaptations that 
help conserve water. Long loops of Henle 
in the kidneys are an adaptation that 
allows for maximum reabsorption of 
water, which is crucial for survival 
in such conditions. This feature is 
likely to be found in animals living in 
arid environments. (B) Storage of water 
in fatty tissues: Some animals have 
evolved to store water in their bodies 
to survive long periods without access 
to water. Storing water in fatty tissues 
is an effective way to ensure a supply 
of water during droughts or dry periods. 
This is another adaptation that would be 
beneficial in a hot arid environment. (C)
Large ears to aid in heat dispersion: In 
hot environments, animals need to 
dissipate excess body heat to avoid 
overheating. Large ears can help with this
by increasing the surface area through 
which heat can be lost. This adaptation 
is also likely to be found in animals 
living in hot arid environments. (D) 
Short loops of Henle to maximize water 
secretion: Short loops of Henle would 
result in less water reabsorption and 
more water being excreted. In a hot arid 
environment where water is scarce, this 
would be a disadvantage because it would 
lead to greater water loss. Therefore, 
this feature is least likely to be found 
in animals living in such conditions. So
the correct answer option is (D) Short 
loops of Henle to maximize water 
secretion

Question: A survey of 1000 registered 
voters revealed that 450 people would vote
for candidate A in an upcoming election. 
If 220,000 people vote in the election, 
how many votes would the survey takers 
predict candidate A should receive?\n(A)
44,500\n(B) 48,900\n(C) 95,000\n(D) 99,000

Answer: To predict the number of votes
candidate A should receive based on the 
survey, we can use a proportion. The 
survey indicates that 450 out of 1000
registered voters would vote for candidate
A. We can set up the following proportion 
to find the predicted number of votes out
of 220,000 voters: \\[\\frac{450}{1000} = 
\\frac{x}{220,000}\\] To solve for \\( x 
\\), we can cross-multiply and then divide:
\\[450 \\times 220,000 = 1000 \\times x\\] 
\\[99,000,000 = 1000x\\] \\[x = 
\\frac{99,000,000}{1000}\\] 
\\[x = 99,000\\] Therefore, the survey
takers would predict that candidate A 
should receive \\( 99,000 \\) votes. So 
the correct answer option is (D) 99,000

Question: A cannonball is fired 
horizontally at 10 m/s from a cliff. Its
speed one second after being fired is 
about\n(A) 10 m/s\n(B) 14 m/s\n(C) 16
m/s\n(D) 20 m/s

Answer: To determine the speed of the 
cannonball one second after being fired 
horizontally, we need to consider both the
horizontal and vertical components of its 
motion. 1. Horizontal Motion: - The 
horizontal velocity (v_x) remains constant
because there is no horizontal acceleration
(assuming air resistance is negligible). -
Given: v_x = 10 m/s 2. Vertical Motion: - 
The vertical velocity (v_y) changes due to 
the acceleration caused by gravity (g = 
9.8 m/s^2). - Initially, the vertical 
velocity is zero because the cannonball 
is fired horizontally. - After one second,
the vertical velocity can be calculated 
using the equation: v_y = g * t where t = 
1 s. v_y = 9.8 m/s^2 * 1 s = 9.8 m/s 3. 
Resultant Speed: - The resultant speed (v) 
is found by combining the horizontal and 
vertical components using the Pythagorean 
theorem: v = sqrt(v_x^2 + v_y^2) v = 
sqrt((10 m/s)^2 + (9.8 m/s)^2) v = sqrt(100
+ 96.04) v = sqrt(196.04) v \approx 14 m/s 
Therefore, the speed of the cannonball one
second after being fired is about 14 m/s. 
So the correct answer option is (B) 14 m/s

Question: A 3-microfarad capacitor is 
connected in series with a 6-microfarad 
capacitor. When a 300-volt potential 
difference is applied across this 
combination, the total energy stored in
the two capacitors is\n(A) 0.09 J\n(B) 
0.18 J\n(C) 0.27 J\n(D) 0.41 J

Answer: To find the total energy stored 
in the two capacitors connected in series,
we need to follow these steps: 1. 
Calculate the equivalent capacitance of
the series combination: For capacitors in 
series, the reciprocal of the equivalent
capacitance \( C_{\text{eq}} \) is the 
sum of the reciprocals of the individual 
capacitances: \[ \frac{1}{C_{\text{eq}}} 
= \frac{1}{C_1} + \frac{1}{C_2} \] Given 
\( C_1 = 3 \, \mu\text{F} \) and \( C_2 = 
6 \, \mu\text{F} \): \[ \frac{1}
{C_{\text{eq}}} = \frac{1}{3 \, 
\mu\text{F}} + \frac{1}{6 \, \mu\text{F}}
= \frac{2}{6 \, \mu\text{F}} + \frac{1}{6
\, \mu\text{F}} = \frac{3}{6 \, 
\mu\text{F}} = \frac{1}{2 \, \mu\text{F}}
\] Therefore, \[ C_{\text{eq}} = 2 \, 
\mu\text{F} \] 2. Calculate the total 
energy stored in the equivalent capacitor: 
The energy \( E \) stored in a capacitor 
is given by: \[ E = \frac{1}{2} C V^2 \]
Here, \( C = C_{\text{eq}} = 2 \, 
\mu\text{F} = 2 \times 10^{-6} \, \text{F}
\) and \( V = 300 \, \text{V} \): \[ E =
\frac{1}{2} \times 2 \times 10^{-6} \, 
\text{F} \times (300 \, \text{V})^2 \] \[
E = \frac{1}{2} \times 2 \times 10^{-6} 
\times 90000 \] \[ E = 1 \times 10^{-6} 
\times 90000 \] \[ E = 0.09 \, \text{J} 
\] Therefore, the total energy stored in 
the two capacitors is \( \boxed{0.09 \, 
\text{J}} \). So the correct answer option
is (A) 0.09

\end{verbatim}



\subsection{Instruction-Tuned Model Experiments}
\label{sec:instruct_experiments}

To investigate how our approach affects models of different training stages, we also include experiments show that our framework also benefits the LLaMA-3-8B-Instruct version in Figure \ref{fig:main_instruct} and Figure \ref{fig:mmlu_instruct}, which has undergone general instruction-tuning \cite{llama2} to enhance performance in many aspects.
Notably, we observe improvements on both the base and the instruction-tuned model versions, which suggests that our approach may generalize well even to well-trained models.

\subsection{Commonsense and General Reasoning}
\label{sec:csqa}

We additionally study the performance of our method on commonsense and general reasoning tasks, specifically evaluating it on CommonsenseQA \cite{talmor-etal-2019-commonsenseqa}, Winogrande \cite{Sakaguchi2019WinoGrande}, and the full MMLU \cite{hendrycks2021measuringmmlu} dataset as presented in Table \ref{tab:csqa}. Despite the strong SFT baseline, which suffices for most questions requiring only one or two reasoning steps, we demonstrate that our method surpasses other preference optimization methods in terms of multi-step reasoning when the model is prompted to think step-by-step. Notably, on the Winogrande dataset, our framework achieves a significant improvement of 6.6\% over the strongest preference optimization baseline, ORPO.

\begin{table}[!t]
    \centering
    \resizebox{1.0\linewidth}{!}{
    \begin{tabular}{lccccccc}
    \toprule
    \textbf{Method}
    &\textbf{CSQA} 
    &\textbf{Winogrande} 
    &\textbf{MMLU} \\
    \midrule
    \textbf{LLaMA-3-8B} \\
    \enspace SFT &  82.7 & 86.0 & 63.7 \\
    \enspace RFT &  72.3 & 64.2 & 59.6 \\
    \enspace DPO &  72.7 & 57.9 & 56.5 \\
    \enspace ORPO & 76.8 & 67.0 & 59.7 \\
    \rowcolor{lightgray} \enspace Ours & 79.7 & 73.6 & 62.0 \\
    \midrule
    \textbf{LLaMA-3-8B-Instruct} \\
    \enspace SFT &  82.3 & 84.8 & 63.4 \\
    \enspace RFT &  73.8 & 66.5 & 63.3 \\
    \enspace DPO &  70.4 & 63.7 & 67.9 \\
    \enspace ORPO & 74.0 & 67.7 & 63.9 \\
    \rowcolor{lightgray} \enspace Ours & 77.7 & 71.2 & 65.6 \\
    \bottomrule
    \end{tabular}
    }
    \caption{Additional evaluation results on commonsense and general reasoning tasks.}
    \label{tab:csqa}
\end{table}

\subsection{Evaluation of Reasoning Quality}
\label{sec:inter_f1}

To quantitatively measure the reasoning quality after training with different methods, we report the Inter. F1 metrics \cite{wang-etal-2023-towardsinterf1} which compares the numerical objects that are consistent between the generated reasoning path and ground-truth reasoning path.
We report the results for LLaMA-3-8B on GSM8K as shown in Table \ref{tab:inter_f1}.
The results demonstrate that our framework not only improves the final reasoning benchmark score, but also enhances the reasoning quality as measure by the Inter. F1 metric.

\end{document}